\documentclass[pdflatex,sn-nature]{sn-jnl}
\usepackage{graphicx}%
\usepackage{multirow}%
\usepackage{amsmath,amssymb,amsfonts}%
\usepackage{amsthm}%
\usepackage{mathrsfs}%
\usepackage[title]{appendix}%
\usepackage{xcolor}%
\usepackage{textcomp}%
\usepackage{manyfoot}%
\usepackage{booktabs}%
\usepackage{algorithmicx}%
\usepackage{algpseudocode}%
\usepackage{listings}%
\usepackage{hyperref}

\usepackage{amsmath,amssymb,amsfonts}
\usepackage{graphicx}
\usepackage{textcomp}

\usepackage{subfig}

\usepackage{algpseudocode}
\usepackage{algorithm2e}

\usepackage{amsmath}
\usepackage{array}

\usepackage{amssymb}
\usepackage{graphicx}

\usepackage{url}
\usepackage{placeins}

\usepackage{pifont}% http://ctan.org/pkg/pifont
\newcommand{\cmark}{\ding{51}}%
\newcommand{\xmark}{\ding{55}}%

\newcolumntype{C}[1]{>{\centering\let\newline\\\arraybackslash\hspace{0pt}}m{#1}}

\usepackage{placeins}

\begin{document}

\title{Can machines reason non-verbally?}

\author
{Shaheer U. Saeed$^{1,2,3,\ast}$,  Zion Tse$^{1,2}$, Gregory Slabaugh$^{1}$,\\ Daniel C. Alexander$^{3}$, Yipeng Hu$^{3}$\\
\\
\normalsize{$^{1}$Digital Environment Research Institute, Queen Mary University of London}\\
\normalsize{$^{2}$Centre for Bioengineering, Queen Mary University of London}\\
\normalsize{$^{3}$UCL Hawkes Institute, University College London}\\
\normalsize{$^\ast$Correspondence e-mail: shaheer.saeed@qmul.ac.uk}
}

\abstract{
Modern machines can solve equations, follow logic and reason through language, yet they still struggle with problems that require visual non-verbal reasoning. This is evident even in advanced systems such as DeepSeek, ChatGPT and Gemini, which remain limited on visual tasks such as completing abstract patterns or delineating ambiguous regions-of-interest. Unlike explicit intermediate steps in verbal reasoning which can be verified through logic or automated tests, visual reasoning involves implicit intermediate steps with no clear methods for verification. Non-verbal reasoning is therefore challenging to model and, as a result, remains a fundamental limitation of current machine intelligence.
Here we introduce GenVerOpt, a general framework that unifies generation, verification and optimisation into a reasoning pipeline for machines. By generating candidate solutions, verifying their correctness, and iteratively refining them, GenVerOpt enables per-sample optimisation rather than relying solely on dataset-level trends captured during model training. This process mirrors human problem-solving where multiple possibilities are imagined, evaluated, and refined until a correct solution emerges.
Across diverse domains, GenVerOpt demonstrates capabilities beyond existing systems: it surpasses humans on abstract IQ puzzles, matches clinicians in prostate cancer segmentation, exceeds experts in liver vessel reconstruction and outperforms state-of-the-art approaches in tumour boundary alignment. 
These results show that machine reasoning can extend beyond language, establishing visual non-verbal reasoning as a foundation for more flexible and generalisable intelligence.
}


\maketitle

\subsection*{The non-verbal roots of human reasoning}
Non-verbal reasoning: the capacity to generate, evaluate and iteratively refine solutions without the use of language, is a fundamental component of intelligence \cite{heyes2012new, gowlett2016discovery, evans1984heuristic, kahneman20122}. 
In early human evolution, this ability was crucial for survival, such as for detecting camouflaged predators, tracking prey, and interpreting complex visual cues \cite{sherwood2008natural}. The same cognitive abilities later enabled more abstract reasoning, such as linking friction to fire-making or crafting tools based on the observation of material interactions \cite{gowlett2016discovery}. Crucially, these abilities predated language and cannot be reduced to verbal thought \cite{gowlett2016discovery}. Although language has now become a powerful tool for articulating and extending reasoning, it also introduces biases tied to specific linguistic structures \cite{wolff2011linguistic}. This raises the question of whether language is best viewed as a complement to, or simply an extension of, more general non-verbal reasoning \cite{kahneman2009conditions, bilalic2008good}. Evidence from human cognitive development further supports this distinction: children reason about space, numeracy and causality long before acquiring language \cite{baillargeon2004infants,baillargeon2008innate}. These insights suggest that non-verbal reasoning operates independently, underscoring its primacy and indispensability in intelligent thought.

\subsection*{The gap between verbal and non-verbal reasoning in machines}
Despite the non-verbal origins of human reasoning, recent advances in machine reasoning have focused predominantly on language-based tasks \cite{deepseek_r1, wei2022chain}. Advances in natural language processing (NLP) have enabled models to solve algebraic problems, perform logical inference, and generate code, capabilities that rely on structured verbal reasoning \cite{deepseek_r1, wei2022chain, kahneman20122}. 
Reasoning has also been successfully modelled in structured domains such as board games, where search and optimisation can identify effective sequences of actions \cite{feng2023alphazero,anthony2017thinking}.
This progress has been driven in part by the availability of well-defined verification methods, such as formal logic rules, unit tests and game outcomes, that make it possible to check whether a proposed solution is correct. This verifiability allows models to attempt multiple intermediate steps, discard those that fail, and iteratively refine these steps until a valid solution is reached. In this way, reasoning in machines differs from simple pattern recognition, where it is not limited to mapping inputs to outputs, but instead involves generating, verifying, and optimising candidate solutions under a clear criterion of correctness.

\begin{figure}[!ht]
    \centering
    \includegraphics[width=0.98\textwidth]{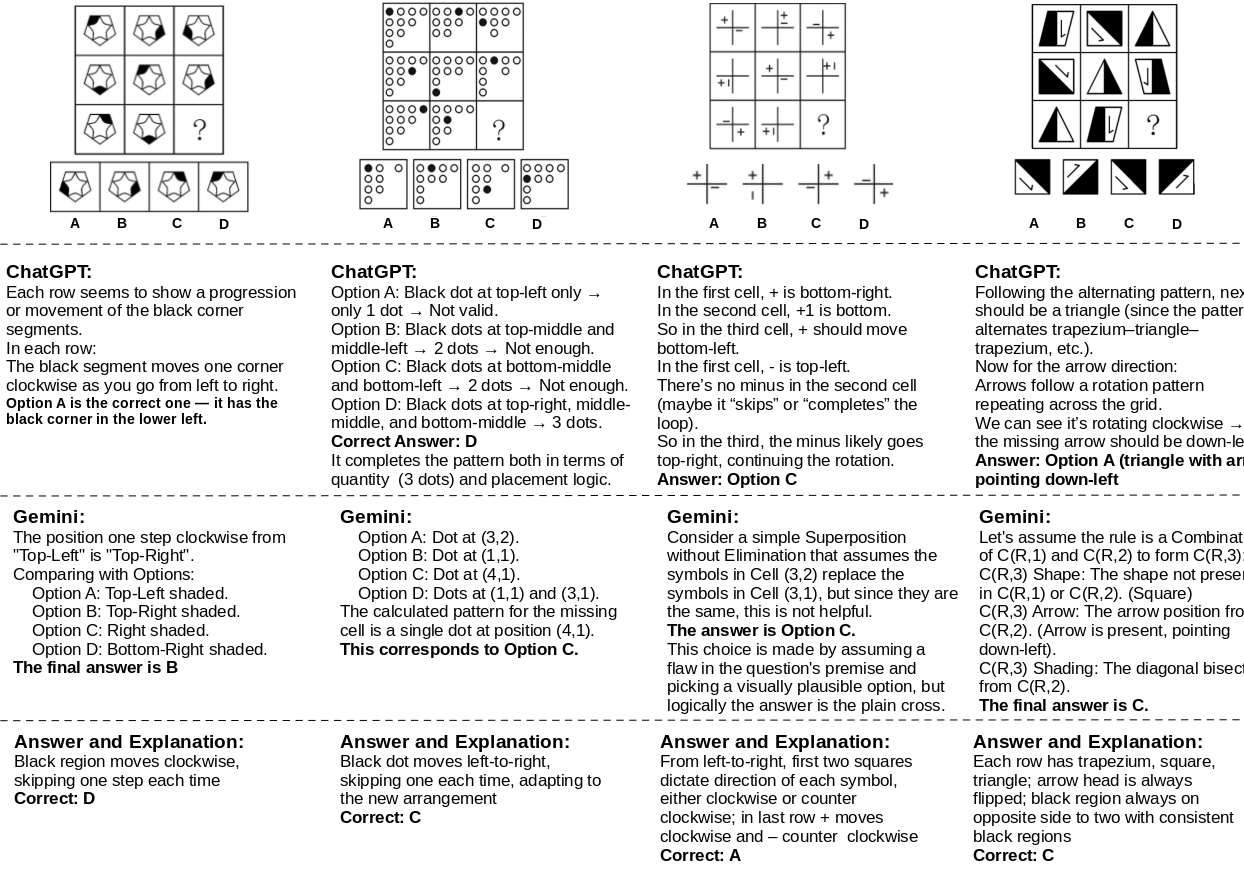}
    \caption{IQ test puzzles solved by recent state-of-the-art reasoning AI models.}
    \label{fig:reasoning_examples}
\end{figure}

However, reasoning approaches developed for language-based or structured domains do not readily extend to visual tasks, even in state-of-the-art multimodal systems \cite{murlidaran2021comparing, cai2025mm, shojaee2025illusion, wolf2025your}. Unlike language or board games, visual reasoning often involves implicit, non-discrete intermediate steps and lacks clear mechanisms for verification. Moreover, some tasks, such as tracing the continuation of a visual pattern, are solved more naturally through direct visual inference than by translating them into words, where verbalisation is typically a post hoc description rather than the reasoning process itself \cite{kahneman2009conditions, gobet2012concepts, bilalic2008inflexibility}.
As illustrated in Fig. \ref{fig:reasoning_examples}, advanced systems such as ChatGPT and Gemini, which combine text and image processing, struggle with tasks that require spatial inference, abstract pattern recognition, or multi-step visual problem-solving. Notably, this limitation persists even when visual reasoning problems are reduced to multiple-choice tasks with a small set of candidate solutions and an explicitly verifiable correct answer. These limitations may stem from reasoning pipelines in multimodal systems being optimised for structured, verifiable tasks in language, rather than for the implicit forms of inference demanded by vision \cite{mitchell2025artificial, cai2025mm}. While specialised domain-specific approaches exist for particular visual tasks \cite{pan2025dinor1, chen2025fast, saeed2025reasoning}, general-purpose visual reasoning remains an unsolved challenge. 
This persistent gap between verbal and visual reasoning highlights a critical limitation in current machine intelligence.

A key barrier to extending reasoning to more complex visual problems is the lack of reliable rules or automated mechanisms for verifying solutions. Unlike multiple-choice puzzles, many real-world visual tasks have high-dimensional solution spaces for which correctness cannot be specified through a predefined answer or simple rule. Verification instead often depends on human interpretation of complex image data.
While human-labelled datasets can provide supervision, many reasoning tasks are too complex to capture with sufficiently large labelled-datasets, with the additional constraint of high annotation costs in many specialised domains \cite{wang2024comprehensive}. Furthermore, intermediate visual reasoning steps are often abstract and not directly representable by humans, making them difficult to formulate into supervision signals for machines.
Intelligence quotient (IQ) puzzles (see Fig. \ref{fig:reasoning_examples}) therefore provide a controlled example of the visual reasoning problem where constraining the task to multiple-choice answers substantially reduces the solution space and makes verification explicit, yet state-of-the-art multi-modal and task-specific models still perform near chance (approximately 25\% average accuracy), far below human performance (56\% accuracy) \cite{cai2025mm}.

\subsection*{Bridging the machine reasoning gap}
Developing models capable of reasoning beyond language could unlock new frontiers in machine intelligence, particularly in tasks that require, or are more effectively solved by, visual reasoning. Beyond the abstract IQ puzzles described earlier, medical image analysis offers concrete examples where success depends not just on recognition but on multi-step reasoning about structure and context. 
In prostate cancer segmentation on MR images, for example, tumour boundaries are often ambiguous, requiring integration of conflicting visual cues across slices and anatomical landmarks to arrive at a plausible delineation \cite{promis}.
In liver vessel segmentation on CT images, spatial reasoning is essential: a human observer can recognise when a broken vessel segment should be linked to its parent branch, or when two similar-looking structures actually belong to different trees, both decisions that involve chaining relationships rather than simple pixel-level classification \cite{chierici2024vascular}. Similarly, in image registration, aligning pre-operative MR with intra-operative ultrasound requires reasoning about how tissue has deformed, inferring correspondences that go beyond local appearance \cite{huang2024one}.
Current data-dependent deep learning models fail to capture such intricacies because curating a sufficiently representative dataset for such problems is challenging due to high variability not only in the problems themselves but also in the expert annotations due to disagreements.
These examples highlight a fundamental distinction: while current deep learning models excel at direct mapping from input to output, they lack the capacity to perform multi-step, non-verbal reasoning needed to resolve ambiguity, enforce structural consistency, and generalise beyond training data.

\FloatBarrier

\newpage

\section*{A recipe for reasoning - Generate, Verify, Optimise}

Psychology literature has identified use of extended thinking time to be one of the main mechanisms for human reasoning \cite{kahneman2003maps, kahneman20122, evans2003two}. Thinking time can be consumed to consider multiple decision strategies and their outcomes in greater detail, leading to improved solutions. This process involves `imagining' potential solutions, deciding how likely each imagined solution is to be successful, and then selecting the solution with the highest success likelihood. 

In machines, reasoning processes can be modelled in a similar framework, involving: 1) generation: which produces a variety of candidate solutions; 2) verification: which determines the correctness likelihood of each solution; 3) optimisation: which allows selection of the most optimal solution. We call this the Generation - Verification - Optimisation, or GenVerOpt, framework.

The generation of candidate solutions may be sourced from previous experience across diverse tasks, sampled from a solution space, or acquired from an intelligent agent.
In this sense, generation resembles inference in conventional machine learning, mapping inputs to outputs. The key distinction is that inference typically yields a single deterministic prediction, whereas reasoning requires generating multiple plausible candidates, often sampled stochastically. However, any such generation is likely to be non-exhaustive and potentially erroneous, at least to a certain degree, due to large solution spaces in modern vision tasks. Therefore, the generated solutions need verification, to determine their correctness and optimisation to select the most correct solution. In humans, these three processes are thought to be intertwined. When we consider a problem and refine the specifics of certain decisions, solution generation or `imagination' is guided towards higher likelihoods of correctness. In other words, the generation of candidate solutions is rarely random and is almost always guided towards optimality, even during the generation process. 

% ** add comparison briefly to other things like mcts alhpazero etc and to our prior work; prior work is a specific case implemented for segmentation and requires a very specific formulation to alter the generation by applying transformations to the outputs; here generation is directly optimised in the optimisation.

\begin{figure}
    \centering
    \includegraphics[width=0.8\textwidth]{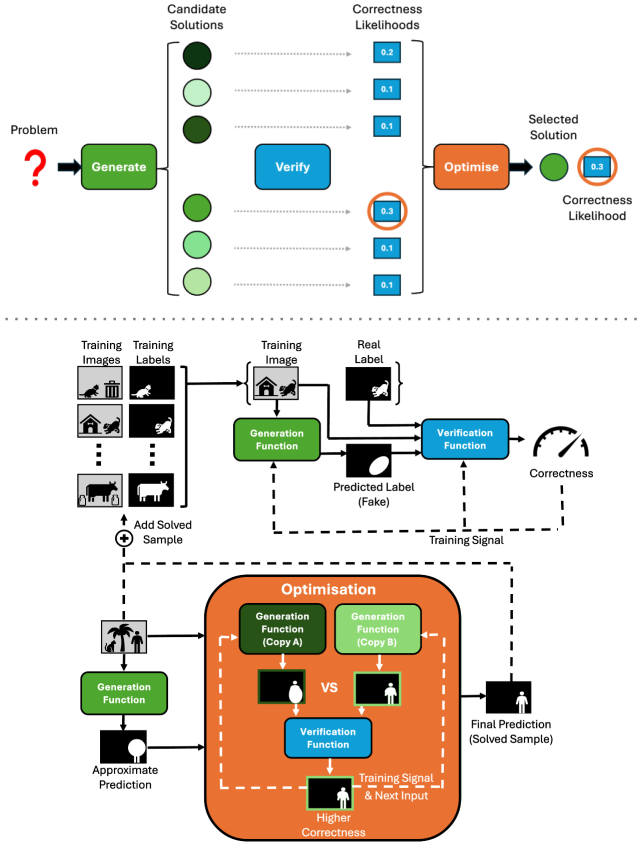}
    \caption{Overview of GenVerOpt (top: conceptual summary; bottom: details of one implementation of GenVerOpt for the dummy task of human segmentation on natural images).}
    \label{fig:genveropt}
\end{figure}

Fig. \ref{fig:genveropt} presents an overview of the general GenVerOpt framework alongside a specific implementation (further details in the `Machine visual reasoning in action' section).

A full mathematical description of the GenVerOpt framework is presented in the `methodological description' section of the supplementary materials .

\subsection*{How GenVerOpt compares to conventional deep learning}

The reasons for modelling GenVerOpt for machine reasoning are twofold. Firstly, verification is easier than generation due to a comparatively lower-dimensional solution space. This is supported by psychology literature, which illustrates the difficulty of generation with an example i.e., generating a good piece of writing is much more difficult than recognising which piece of writing is good or bad \cite{yun2013exploring}. In machine learning literature, similar conclusions have been reached where lower-dimensional tasks such as classification are easy to generalise to, compared to higher dimensional tasks such as segmentation or localisation \cite{nichol2018reptile}. The second reason relates to the optimisation in GenVerOpt compared to conventional deep learning. In conventional deep learning, model weights are optimised with respect to a dataset and the model then infers on unseen samples. This makes models likely to be stuck in local optima, shaped by dataset biases, label subjectivity, or limited sample diversity. In contrast, GenVerOpt performs optimisation at the level of individual samples, where solutions are iteratively proposed and verified for each new input. While not exhaustive, this process allows exploration of multiple plausible solutions per case, for example, alternative boundary placements in ambiguous segmentations, before the final solution, making it less likely to be stuck in local optima caused by dataset-level trends \cite{ahn2022mitigating, guo2019spottune, kohl2018probabilistic}. 
This means that dataset-specific local optima in deep learning, often caused by data biases, subjectivity in human labels and limited data diversity, are unlikely to propagate into the sample-specific optimisation in GenVerOpt.

\section*{Machine visual reasoning in action}

To illustrate the potential of machine visual reasoning, we apply different versions of the GenVerOpt framework to four representative tasks. Each case study implements distinct approaches to generation, verification, and optimisation, reflecting the flexibility of the framework.

\subsection*{Abstract visual reasoning: IQ test pattern puzzles} 

Visual IQ tests present pattern puzzles in which objects move or transform according to a hidden rule across a grid or sequence. The task is to select the candidate that completes the pattern, requiring abstract reasoning beyond recognition or memorisation. Such puzzles, widely used to assess human intelligence, provide a natural benchmark for evaluating machine reasoning.

\subsubsection*{GenVerOpt implementation}

In this setting, solution generation is trivial: all candidate answers are predefined, and the correct one is guaranteed to be among them. This deliberately constrains the solution space, providing a controlled setting in which visual verification can be isolated from the complexity of solution generation. The challenge lies in verification, judging which option correctly completes the pattern.

To model verification, we train a neural network to evaluate the correctness of each candidate solution. For training, candidates are inserted into the missing square of the puzzle, creating complete grids, labelled as correct or incorrect. The verifier is trained across diverse puzzle types, spanning geometric transformations, spatial relationships, and logical patterns \cite{cai2025mm}. To encourage generalisation to new puzzle types, we adopt the Reptile meta-learning algorithm \cite{nichol2018reptile}. 

The trained network is then fine-tuned using a small, task-specific set of 32 temporal movement puzzles, examples requiring pattern continuation in space and time (see Fig. \ref{fig:reasoning_examples}). During inference, the network outputs a prediction score for each candidate, which we interpret as the likelihood of correctness.

Since the solution space is small and fully enumerated, optimisation reduces to an exhaustive search: the system evaluates all candidates using the verification network and selects the one with the highest predicted correctness. This provides a minimal implementation of GenVerOpt in which generation and optimisation are constrained, allowing the learned verifier to be evaluated directly.

\begin{figure}[!ht]
    \centering
    \includegraphics[width=0.95\linewidth]{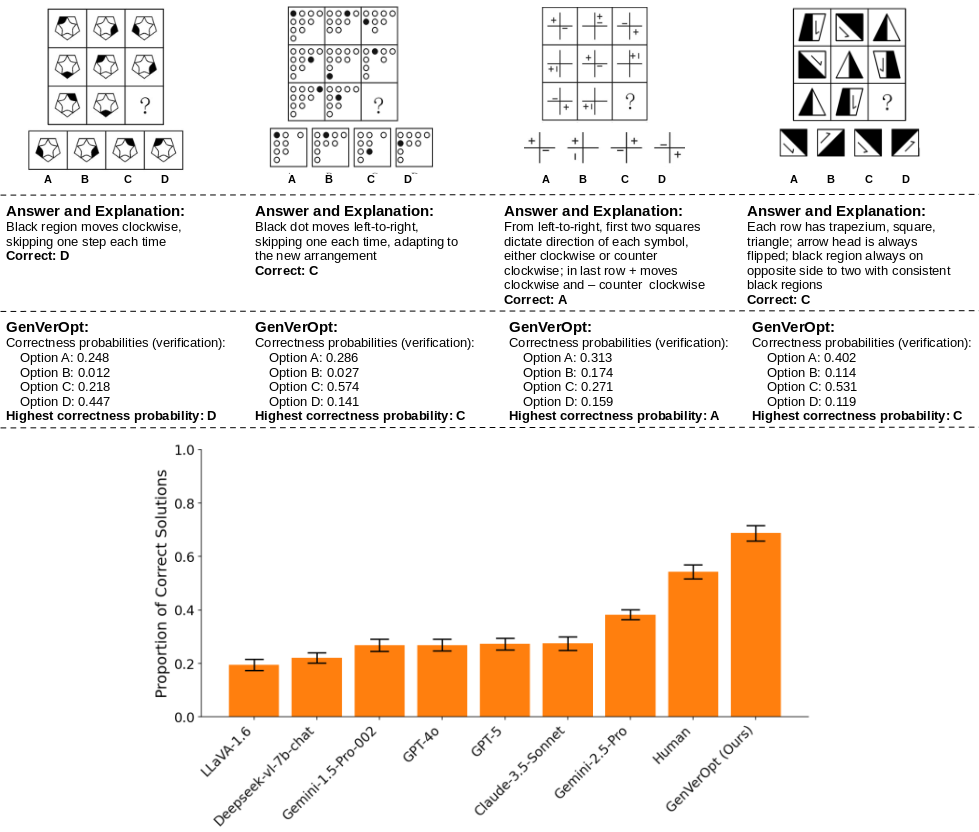}
    \caption{IQ test puzzles solved by GenVerOpt and a comparison with humans and commonly used AI models. Open-source models LLaVA and Deepseek were fine-tuned using low-rank adaptation on the dataset before inference and closed-source models were primed using four-shot prompting with examples randomly selected from the datasets and performance averaged across 8 runs. Task-specific models were trained from scratch on the same data as GenVerOpt.}
    \label{fig:iq_performance}
\end{figure}

\subsubsection*{Surpassing human performance on abstract visual reasoning}

As shown in Fig. \ref{fig:iq_performance}, GenVerOpt outperforms all tested machine reasoning models, and even human participants, on 379 temporal movement puzzles. While existing models perform near the level of chance (25\% accuracy) and fail to capture abstract visual concepts such as skipping, ordering, or continuity (Figs. \ref{fig:reasoning_examples}, \ref{fig:iq_performance}), GenVerOpt consistently identifies the correct answer.

Crucially, its outputs are not binary but graded: ranked probabilities reveal the semantic plausibility of alternatives. For example, in the left panel of Fig. \ref{fig:iq_performance}, clearly incorrect candidates receive the lowest confidence, while options spatially adjacent to the correct solution receive higher scores, that are still lower than the true answer. This behaviour, consistently observed across the dataset, suggests that GenVerOpt learns structural relationships underlying the puzzles, rather than over-fitting to superficial cues.

These results show that even a minimal implementation of GenVerOpt supports robust abstract reasoning, setting a new benchmark for machine performance on visual IQ tests and suggesting a path toward more general non-verbal intelligence.

\FloatBarrier

\begin{figure}
    \centering
    \includegraphics[width=0.95\linewidth]{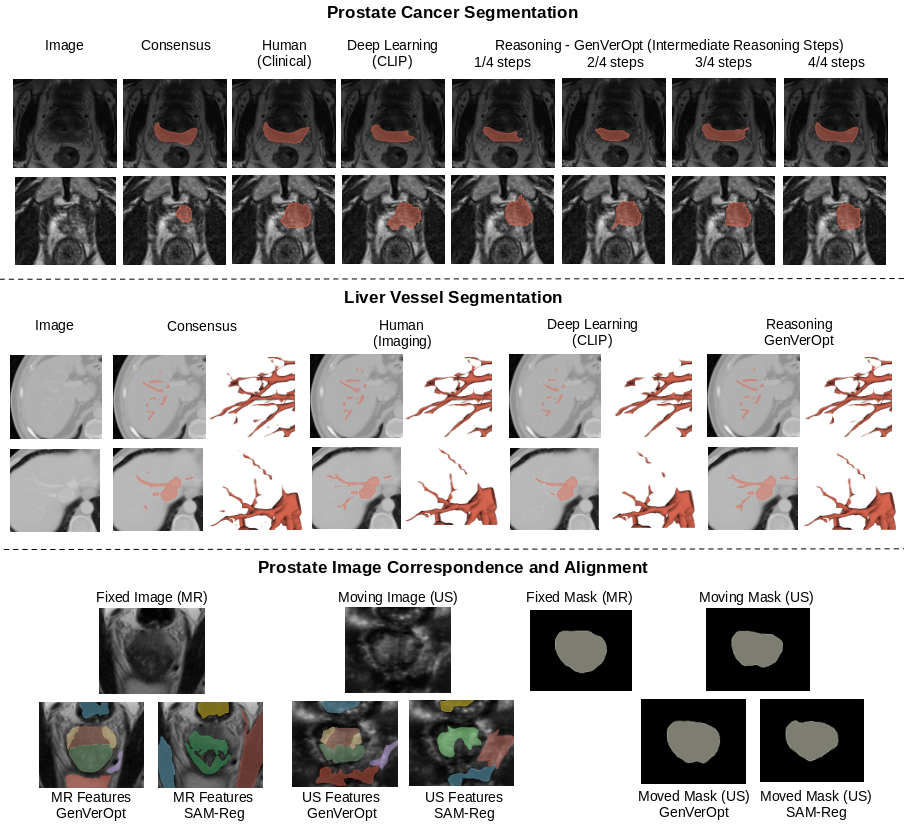}
    \caption{Examples of medical imaging problems solved by GenVerOpt and compared with human experts and recent state-of-the-art AI models.}
    \label{fig:med_performance_samples}
\end{figure}

\subsection*{Resolving ambiguity: prostate cancer segmentation}

Prostate cancer segmentation on MR images involves delineating diseased regions that often appear subtly and inconsistently across patients and sequences \cite{promis}. Ambiguous boundaries, anatomical variability, and low tissue contrast make the task inherently uncertain, even for experts. Accurate segmentation therefore requires reasoning over spatial context and structural cues, providing a strong test of machine reasoning under visual ambiguity.

\subsubsection*{GenVerOpt implementation}

Unlike IQ puzzles, where correctness is binary, medical segmentation involves ambiguous solution spaces with multiple near-correct answers. To capture this, we model generation and verification jointly, allowing both components to co-evolve in response to ambiguity.

This interaction is framed following a framework similar to Goodfellow et al. \cite{goodfellow2020generative}: a generator proposes segmentations, while a verifier predicts their likelihood of correctness.
During training, expert or consensus-labelled segmentations are treated as fully correct (probability 1), while the generator’s outputs are treated as incorrect (probability 0). The generator is trained to maximise correctness measured by verifier scores, while the verifier learns to distinguish fine-grained correctness. Meta-learning with Reptile \cite{nichol2018reptile, nichol2018first} across multiple segmentation tasks enables generalisation across domains.

For a new task, both components are fine-tuned with only 64 labelled task-specific examples. Optimisation is then carried out using reinforcement learning (RL) self-play, where two copies of the generator propose candidate solutions, the verifier evaluates them, and the higher-scoring output is rewarded while the other is penalised. Note that the solutions are sampled probabilistically using the logits (raw probability without thresholding) and thus the two solutions are likely to be unique (further details described in the `methodological description' section of the supplementary materials). The generator is trained, using this reward/ penalty signal, to produce better solutions on the next iterations. Iterating this process refines solutions toward higher verifier correctness. Pairwise comparison improves stability and reduces verifier bias, akin to preference-based reinforcement learning in language models \cite{rafailov2023direct}. Once the verifier scores stabilise, the solution is regarded as complete and the optimisation can be terminated.

While components still incorporate human supervision, such as using labelled examples to guide generation or verification, training on related tasks improves generalisability. Crucially, the use of randomness during solution generation ensures a broad search over plausible candidates, enabling the system to optimise for each individual input. This diversity reduces the risk of dataset-level biases and allows GenVerOpt to produce well-reasoned solutions that go beyond the scope of its training data.

This iterative optimisation captures several key attributes of human-like reasoning \cite{kahneman20122}. First, performance improves with extended `thinking time', as optimisation steps yield predictions with higher verifier correctness, mimicking how humans deliberate before decision-making. Second, the process produces intermediate steps where each iteration provides a candidate solution and its correctness estimate, analogous to human reasoning trails. Finally, the framework demonstrates strong generalisation from minimal supervision, requiring only a small number of labelled samples per task. These properties are further explored in the `supplementary results' section of the supplementary materials.

Together, these components allow GenVerOpt to navigate ambiguity, generalise across medical segmentation tasks, and optimise solutions in a way that mimics the flexibility and deliberation of expert human reasoning.

\subsubsection*{Resolving ambiguous pathology like a trained clinical expert}

As shown in Fig. \ref{fig:med_performance}, GenVerOpt substantially outperforms state-of-the-art automated methods for prostate cancer segmentation, achieving statistically significant gains. Unlike conventional deep learning systems that fail in the face of subtle or conflicting cues, GenVerOpt’s iterative generation–verification cycle allows it to refine solutions until uncertainty is resolved. This is demonstrated in the top example, where although the deep learning and GenVerOpt solutions are initially similar, GenVerOpt refines the tumour boundaries despite a similar appearance of the healthy tissue and tumour.

Strikingly, with only 64 task-specific training examples, GenVerOpt exceeds the accuracy of a human imaging expert while performing at the level of a trained clinician. This demonstrates that reasoning-driven optimisation can overcome the very ambiguities that limit both current algorithms and human–machine parity. By reducing reliance on scarce expert annotations while delivering expert-level accuracy, GenVerOpt offers direct value for clinical workflows, where precise tumour delineation is critical for diagnosis, therapy planning, and patient outcomes.

\subsection*{Inferring structure: outlining liver vessel trees}

Liver vessel segmentation on CT images requires tracing highly variable and branching vascular structures \cite{chierici2024vascular}. Overlapping intensities with surrounding tissue, occlusions, and irregular vessel morphology make the task error-prone, even for experts \cite{wu2023hepatic}. Accurate segmentation requires reasoning over spatial context to link vessels into coherent trees, providing a strong test for structural visual reasoning.

\subsubsection*{GenVerOpt implementation}

For this task, we employ the same generator–verifier framework used in prostate cancer segmentation, fine-tuned with only 64 task-specific vessel annotations. To encourage structural fidelity, we further augment training with examples containing artificially broken vessels, allowing the verifier to penalise discontinuities and guide the generator toward anatomically plausible solutions. Optimisation proceeds as before, iteratively refining predictions until the verifier’s correctness stabilises.

\subsubsection*{Achieving human-like structural awareness}

\begin{figure}
    \centering
    \includegraphics[width=0.97\linewidth]{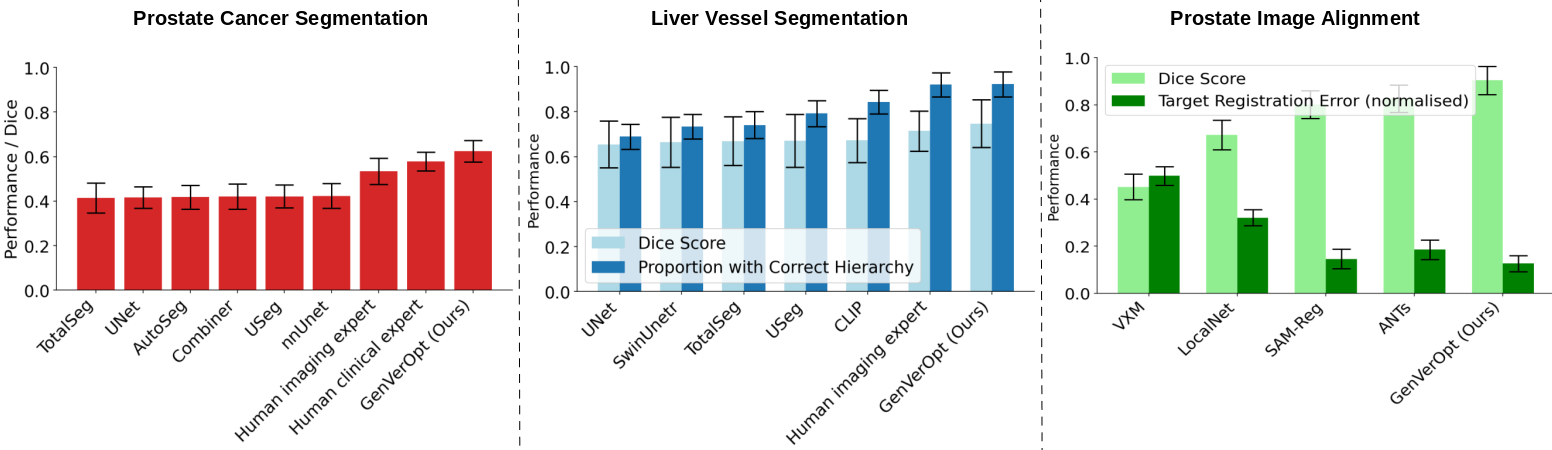}
    \caption{Performance comparison of GenVerOpt against expert humans and recent state-of-the-art AI systems for medical imaging tasks.}
    \label{fig:med_performance}
\end{figure}

As shown in Fig. \ref{fig:med_performance}, GenVerOpt achieves a Dice score on par with a trained human observer while substantially outperforming state-of-the-art automated methods. Crucially, its advantage lies not only in voxel-level accuracy but in reasoning about structure, where GenVerOpt reconstructs vessel hierarchies with a higher proportion of correct parent–child assignments than all baselines, and even slightly surpasses human performance with statistical significance (where human performance is measured as a single expert's segmentation compared with a consensus between three experts used as ground truth). Qualitative results further show that GenVerOpt produces markedly fewer broken or disconnected vessels, demonstrating its ability to enforce global anatomical coherence. Together, these results highlight that GenVerOpt goes beyond segmentation to capture the relational logic of vascular anatomy, a capability with direct implications for surgical planning, interventional navigation, and disease monitoring.

% where's waldo
% train NN gen-ver to detect waldo
% image -> gen -> coord
% coord -> ver -> score (0/1) - train ver on simulated waldo and non-waldo examples
% opt: use rl to refine coord - rl agent is gen

% can discover strategies like searching areas near similar clothed people etc.

% show history of bpounding boxes on a where's waldo image

\subsection*{Identifying correspondence: aligning prostate tumour boundaries}

Prostate tumour alignment between pre-operative MR and intra-operative ultrasound (US) is complicated by tissue deformation, variability in tumour appearance, and partial occlusions during surgery \cite{baum2023mr, huang2024one}. Accurate alignment requires reasoning beyond local appearance, integrating spatial and structural cues to infer the most plausible correspondence between boundaries, critical for ensuring accurate biopsies and effective focal treatments.

\subsubsection*{GenVerOpt implementation}

In this task, GenVerOpt predicts corresponding regions-of-interest across MR and US images. Solution generation is built on the Segment Anything Model (SAM) \cite{kirillov2023segment}, with prompts controlled by a prompt-predicting neural network that outputs paired locations across both modalities. Segmentations are obtained using positive-only prompts, applied one at a time on US and MR respectively. Each image pair is sampled with 16 corresponding prompts in total, yielding corresponding segmentations on both images. These corresponding segmentations are then used to compute a dense displacement field that warps the US image into alignment with the MR image.

Verification is based on two complementary measures: the Dice overlap between the SAM-segmented regions post-registration and the structural similarity index (SSIM) between the warped and fixed images. These scores quantify both anatomical correspondence and image-level consistency.

Optimisation uses RL to train the prompt-predicting neural network to predict prompt locations, given an input pair of images. Each candidate set of points is rewarded according to the verification score, guiding the system toward prompts that maximise alignment accuracy. By learning to predict point locations that jointly optimise Dice and SSIM, GenVerOpt discovers correspondences that enable efficient and robust registration.

Performance is evaluated using target registration error (TRE) on unseen landmarks (normalised by assuming unaligned 12.5mm as the maximum error), as well as Dice overlap on anatomical structures such as the prostate gland, which are separate from the SAM-segmented regions in the generation and verification steps. These external metrics ensure that alignment quality reflects genuine generalisation rather than over-fitting to supervision.

\subsubsection*{Outperforming state-of-the-art}

GenVerOpt achieves statistically significantly higher Dice overlap and lower TRE compared to leading registration methods, reflecting its ability to adapt flexibly to patient-specific deformations and appearance shifts. By delivering more reliable tumour alignment, GenVerOpt has the potential to improve intra-operative guidance and directly enhance the precision of prostate cancer interventions.

\section*{Non-verbal reasoning: a pathway to more intelligent machines}

%%%%%%

Human intelligence was built on non-verbal reasoning long before language, enabling survival, discovery, and problem-solving in complex environments. Bringing this capacity into machine intelligence, through the GenVerOpt framework, enables progress on a diverse set of challenging visual reasoning tasks, from abstract pattern puzzles to complex problems in medical imaging. 

GenVerOpt demonstrates that reasoning in machines need not rely on language or vast labelled datasets. Instead, iterative cycles of generation, verification, and optimisation can allow systems to deliberate, adapt, and refine their solutions; resolving ambiguity, generalising from minimal supervision, and exposing interpretable reasoning trails.

Our case studies suggest a general pathway for machine intelligence: one where systems can move beyond recognition to true problem-solving, capable of handling uncertainty and structural complexity across domains. Such an approach could transform healthcare, where reasoning about anatomy and pathology is critical, and extend more broadly to scientific discovery and real-world decision-making.

Equipping machines with non-verbal reasoning may be key to achieving flexible intelligence beyond the limits of language and current datasets.

\backmatter

\section*{Acknowledgments}

\paragraph{Funding:} This work was supported by the International Alliance for Cancer Early Detection, an alliance between Cancer Research UK [EDDAPA-2024/100014] \& [C73666/A31378], the University of Cambridge, the OHSU Knight Cancer Institute, the German Cancer Research Center (DKFZ), the University College London, the University of Manchester, and the Dana Farber Cancer Institute; and the National Institute for Health Research University College London Hospitals Biomedical Research Centre.
% ucl data safehaven, isambard

\newpage

\section*{Supplementary materials}

\subsection*{Contents}

Methodological description\\
Supplementary results (ablations)\\
Computational cost\\
Fig. S6 to S7\\
Algo. S1 to S2\\
Tables S1 to S5\\
References 69 to 84\\

%%%%%%%%%%%%%%%%%%%%%%%%%%%%%%%%%%%

\subsection*{Methodological description}
\label{sec:method}

\subsubsection*{GenVerOpt formalisation}
% maybe make some equations numbered in this section

\paragraph{Generation:} Generation produces candidate solutions, given an input. Let $x\in\mathcal{X}$ denote the image and $y\in\mathcal{Y}$ denote the solution, where $\mathcal{X}$ is the domain of images and $\mathcal{Y}$ is the domain of solutions (e.g., the solution could be labels, segmentations or deformation fields). The generation module is a stochastic function $g(\cdot; \theta):\mathcal{X}\rightarrow\mathcal{P}(\mathcal{Y})$, parametrised by $\theta$, where $\mathcal{P}(\mathcal{Y})$ is the space of probability distributions over $\mathcal{Y}$.

For an input $x\in\mathcal{X}$, the generator $g(x; \theta)$ specifies a distribution over possible solutions. Candidate solutions are sampled using $y\sim g(x;\theta)$, where multiple independent draws yield alternative candidates.

\paragraph{Verification:} Verification determines the correctness probability of a given input-solution pair. The verification module is modelled as a function $v(\cdot; \phi): \mathcal{X} \times \mathcal{Y} \rightarrow [0,1]$, parametrised by $\phi$. For an input-solution pair $(x,y)$, the correctness score is given by $v(x,y; \phi)\in[0,1]$, which can be interpreted as the likelihood that $y$ is a correct solution for $x$.

\paragraph{Optimisation:} Optimisation aims to identify, for each input, the most plausible solution according to the verifier. Formally, the solution is given by $y^* = \arg\max_{y\sim g(x;\theta)} ~\mathbb{E}\left[v(x,y; \phi) \right]$. 

In practice, the expectation is approximated by iterative sampling and refinement: the generator proposes candidate solutions, the verifier scores them, and the process is repeated until the solution stabilises. Crucially, this optimisation is performed per sample, allowing the framework to deliberate and adapt to each new input independently of dataset-level supervision.

\subsubsection*{Learning generation and verification before reasoning}

Before inference-time reasoning can take place, the generation and verification modules must be jointly optimised so that they provide meaningful candidates and correctness estimates. This optimisation is performed at the dataset level or across datasets.

While in principle both modules may be non-parametric, as demonstrated in some of our case studies, for generality we describe them here in their parametric form. The generator $g(x;\theta)$ is trained to produce candidate solutions that approach the distribution of correct solutions, while the verifier $v(x,y;\phi)$ is trained to assign high scores to correct input-solution pairs 
$(x,y)$ and low scores otherwise. 

One of the strategies described in our case studies is joint training, where the verifier learns to discriminate between correct and generated solutions, while the generator is updated to maximise verifier scores. This interplay encourages the generator to produce more realistic candidates and the verifier to sharpen its discrimination.

Formally, the joint objective can be expressed as:

\begin{equation}\label{eq:joint_obj}
    \min_\theta \max_{\phi} ~\mathbb{E}_{x\sim\mathcal{X}} \left[  \mathcal{L} (v(x, ~g(x; \theta); ~\phi), ~ c(x,y)) \right]
\end{equation}

where $c(x,y)\in[0,1]$ is the correctness label (1 for expert-labelled solutions, 0 for generated solutions).

In our formulation, $\mathcal{L}(\cdot)$ can be expressed as two complementary losses, one for the verifier and one for the generator.

\paragraph{Verifier objective:} The verifier should approximate $c(x,y)$. Using binary cross-entropy, the verifier objective is framed as:

\begin{equation}\label{eq:verifier_obj}
    \mathcal{L}_v(\phi) = \mathbb{E}_{x,y}\left[  ~v(x,y;\phi)\times \log{c(x,y)} ~ - ~ (1-v(x,y;\phi))\times \log(1-c(x,y))  \right]
\end{equation}

\paragraph{Generator objective:} The generator should maximise the verifier-predicted correctness of the generated solutions. Therefore, the generator objective maximises the verifier score:

\begin{equation}\label{eq:generator_obj}
    \mathcal{L}_g(\theta) = \mathbb{E}_{(x,y)\sim g(x; \theta)} \left[  v(x,y;\phi) \right]
\end{equation}

\paragraph{Alternating optimisation:} The joint objective summarised in Equation \ref{eq:joint_obj} is a proxy for alternating updates of the verifier (Equation \ref{eq:verifier_obj}) using $\min_\phi \mathcal{L}_v$ and the generator (equation \ref{eq:generator_obj}) using $\max_\theta \mathcal{L}_g$.

\paragraph{Cross-task meta-optimisation:} To enable generalisation across tasks, we extend this training with the Reptile meta-learning algorithm. Let $\{\mathcal{D_i}\}_{i=1}^M$ denote datasets from related tasks (e.g., if the target task is cancer segmentation then related tasks can be organ segmentation). For each dataset $\mathcal{D_i}$, the parameters $(\theta, \phi)$ are optimised using a single step of the alternating optimisation above, producing task-specific solutions $(\theta_i,\phi_i)$. The shared cross-task parameters can then be updated as:

\begin{equation}
    (\theta, \phi) \leftarrow (\theta, \phi) + \alpha \frac{1}{M} \sum_{i=1}^M \left[  (\theta_i, \phi_i) -(\theta, \phi) \right]
\end{equation}

where $\alpha$ is the step-size.

The cross-task optimisation encourages the generator and verifier to acquire task-agnostic priors that can be rapidly adapted to new tasks with limited supervision.

\subsubsection*{Reinforcement learning self-play for optimisation}

\paragraph{Reinforcement learning}
One of the mechanisms for optimisation described in our case studies is RL self-play. 

When using simple single-agent RL, the generator $g(x;\theta)$ defines a distribution over candidate solutions and the verifier provides a likelihood of correctness $v(x,y; \phi)$.

The optimisation objective is therefore to maximise the expected verifier score:

\begin{equation}
    J(\theta|x) = \mathbb{E}_{y\sim g(x;\theta)}\left[  v(x,y; \theta) \right]
\end{equation}

The optimisation goal is then:

\begin{equation}
    y^* = \arg \max_{y\sim g(x;\theta)}  \mathbb{E} \left[  J(\theta|x) \right]
\end{equation}

This corresponds to the optimisation goal stated earlier (in the `GenVerOpt formalisation - Optimisation' section).

\paragraph{Self-play}
To estimate the gradients of $J(\theta)$, we use a self-play strategy. Two candidate solutions $y_1, y_2 \sim g(x;\theta)$ are sampled independently and the verifier assigns scores to each $v(x, y_1; \theta)$ and $v(x,y_2; \theta)$. A relative reward can then be computed as $R = v(x, y_1; \theta)-v(x,y_2; \theta)$. 

The generator parameters are updated with a policy-gradient step:
\begin{equation}
\nabla_\theta J(\theta \mid x) \approx \Delta R , \big(\nabla_\theta \log g(y_1 \mid x;\theta) - \nabla_\theta \log g(y_2 \mid x;\theta)\big).
\end{equation}

Using this for updates in gradient ascent increases the probability of sampling higher-scoring candidates and decreases the probability of lower-scoring ones, thereby shifting the generator distribution toward regions of higher verifier correctness. Repeated iterations progressively refine the generator’s outputs, and the best-scoring candidate at convergence is selected as the final solution. The relative formulation inccentivises the direction of improvement rather than simply relying on an absolute score, this ensures that any small fluctuations in absolute discriminator scores are less likely to impact the learning process.

\newpage

\subsection*{Materials and datasets}

\subsubsection*{IQ test pattern puzzles}

For the IQ test puzzles, we utilised the MM-IQ dataset \cite{cai2025mm}. The data used for training and evaluation are summarised in Tab. \ref{tab:iq_data} below.

\begin{table}[!ht]
\centering
\begin{tabular}{|c c c|}
\hline
Dataset & Samples & Role\\
\hline
Logical Operation &  395 & Training\\
Spatial Relationship & 102 & Training\\
Geometric Relationship & 355 & Training\\
\hline
Temporal Movement & 32 + 379 & Fine-tuning + Evaluation\\
\hline
\end{tabular}
\caption{IQ test pattern puzzels dataset}
\label{tab:iq_data}
\end{table}

\subsubsection*{Medical image segmentation}

Both the prostate cancer and liver vessel segmentation tasks utilise the same datasets for the initial training of the verifier and generator. These datasets are summarised in Tab. \ref{tab:seg_datasets} below.

\begin{table}[!ht]
\centering
\begin{tabular}{|c c c c c c|}
\hline
Dataset & Modality & Structure & Cohort Country & Samples & Role\\
\hline
\cite{antonelli_decathalon} & CT & Spleen           & United States     &  41 & Training\\
\cite{synapse_multi_atlas}  & CT & Gallbladder      & United States     & 30 & Training\\
\cite{synapse_multi_atlas}  & CT & Adrenal Gland    & United States     & 30 & Training\\
\cite{synapse_multi_atlas}  & CT & Major Vessels    & United States     & 30 & Training\\
\cite{synapse_multi_atlas}  & CT & Stomach          & United States     & 30 & Training\\
\cite{rister_ct_org}        & CT & Kidneys          & $>$3 countries    & 119 & Training\\
\cite{rister_ct_org}        & CT & Bladder          & $>$3 countries    & 119 & Training\\
\cite{chaos}                & MR & Liver            & Turkey            & 15 & Training\\
\cite{chaos}                & MR & Kidneys          & Turkey            & 15 & Training\\
\cite{chaos}                & MR & Spleen           & Turkey            & 15 & Training\\
% https://chaos.grand-challenge.org/Data/
\cite{amos}                 & MR & Bladder          & China             & 40 & Training\\
\cite{amos}                 & MR & Gallbladder      & China             & 40 & Training\\
\cite{amos}                 & MR & Prostate         & China             & 40 & Training\\
\cite{amos}                 & MR & Major Vessels    & China             & 40 & Training\\
% https://amos22.grand-challenge.org/
\cite{antonelli_decathalon} & MR & Prostate         & Netherlands       & 48 & Training\\
\cite{antonelli_decathalon} & CT  & Liver Tumour & France & 131  & Training\\
\cite{antonelli_decathalon} & CT & Pancreas Tumour & United States & 282  & Training\\
\cite{antonelli_decathalon} & CT & Colon Tumour & United States & 126  & Training\\
\cite{kits} & CT & Kidney Tumour & United States & 489 & Training\\
\hline
\cite{promis} & mp-MR & Prostate Tumour & United Kingdom & 64 + 500  & Fine-tuning + Evaluation\\
\cite{antonelli_decathalon} & CT & Liver Vessels    & United States     & 64 + 200 & Fine-tuning + Evaluation\\
\hline
\end{tabular}
\caption{A summary of the data sets used for training and evaluation for the prostate cancer and liver vessel segmentation tasks.}
\label{tab:seg_datasets}
\end{table}

\subsubsection*{Region correspondence}

The region correspondence task utilised the muRegPro \cite{baum2023mr} dataset which consists of 135 paired MR and US images of the prostate. Segmentations are available for evaluation and include structures such as prostate gland boundaries, zonal structures, water-filled cysts, and calcifications. Landmarks are also identified on tumour boundaries for evaluation.

\newpage

\subsection*{Supplementary results}

All the supplementary results are described for the task of prostate cancer segmentation with performance evaluated on data that was held-out from training.  

% removing components
% varying amount of data used to train
% plot of thinking time with performance
% qualitative analysis of results - chain of thought  - trajectory of samples is needed e.g., just for prostate cancer is enough

\paragraph{Ablating methodological components}

We remove components from our methodological framework described in the `Learning generation and verification before reasoning' and `Reinforcement learning self-play for optimisation' sections. The omission of joint training means that the generator and verifier are trained separately, where the generator is first trained to perform the task using human labels and then a verifier is trained subsequently. The omission of self-play refers to replacing the two copies of the generator in the optimisation process with a single copy and replacing the reward based on difference in verifier scores with an absolute reward based on directly on the single verifier score. The omission of RL refers to replacing RL with joint training as the optimisation process for final reasoning. The results in Tab. \ref{tab:ablation_res_tab} show that all components contributed positively towards performance.

\begin{table}[!ht]
\centering
\begin{tabular}{|C{1.5cm} C{1.5cm} C{1.5cm} | C{1.5cm} C{1.5cm} | C{2.0cm}|}
\hline
\multicolumn{3}{|c|}{Generator + Verifier} & \multicolumn{2}{c|}{Optimiser} &\\
\hline
Joint Training   & Reptile    & Cross-task training  & Self-play  & RL        & Performance \\
\hline
\cmark                 & \cmark     & \cmark        & \cmark             & \cmark   & 62.4$\pm$4.9 \\%
\hline
\cmark                 & \xmark     & \xmark        & \cmark             & \cmark   & 51.4$\pm$5.5 \\ %
\xmark                 & \cmark     & \cmark        & \cmark             & \cmark   & 54.8$\pm$4.6 \\%
\cmark                 & \xmark     & \cmark        & \cmark             & \cmark   & 60.6$\pm$5.4 \\%
\xmark                 & \xmark     & \xmark        & \cmark             & \cmark   & 49.5$\pm$4.7 \\ %
\hline
\cmark                 & \cmark     & \cmark        & \xmark             & \cmark   & 53.8$\pm$5.9 \\%
\cmark                 & \cmark     & \cmark        & \xmark             & \xmark   & 41.7$\pm$5.4 \\ %
\hline
\cmark                 & \xmark     & \xmark        & \xmark             & \xmark   & 32.3 $\pm$6.2 \\ %
\hline
\end{tabular}
\caption{Ablation study - removing components from the framework to test efficacy of different approaches. The standard deviation is across different runs rather than patient cases.}
\label{tab:ablation_res_tab}
\end{table}

\paragraph{Varying amounts of task-specific data before reasoning}

Before initiating the optimisation process in the GenVerOpt framework, the generator and verifier are fine-tuned with task-specific data. The variation in performance with increasing amounts of task-specific data is summarised in Tab. \ref{tab:vary_data}. The trend show an improvement with increasing data up to a plateau at 64 samples. The reasoning or optimisation time per sample decreases as more samples are added with a plateau observed after 64 samples.

\begin{table}[!ht]
\centering
\begin{tabular}{|c c c|}
\hline
Samples & Performance & Reasoning Time / h\\
\hline
16& 56.1$\pm$4.7 & 40\\
32& 59.6$\pm$5.4& 35\\
48& 60.1$\pm$4.9& 31\\
64& 62.4$\pm$4.9 & 26\\
80& 61.5$\pm$5.1 & 24\\
96& 61.7$\pm$5.5 & 25\\
\hline
\end{tabular}
\caption{Measuring performance of GenVerOpt, given different amounts of labelled task-specific examples.}
\label{tab:vary_data}
\end{table}

\paragraph{The relationship between reasoning time and performance}

For 64 task-specific samples used for initiating the optimisation in GenVerOpt, we evaluate how the performance varies with reasoning time i.e., time spent in the RL self-play optimisation. The performance is averaged across 24 samples.

\begin{table}[!ht]
\centering
\begin{tabular}{|c c|}
\hline
Reasoning Time / h & Performance\\
\hline
0 & 38.2$\pm$4.9 \\
4 & 42.4 $\pm$4.9\\
8 & 47.6$\pm$5.2\\
12 & 52.9 $\pm$4.3\\
16 &  55.7$\pm$5.3\\
20 &  58.3$\pm$5.7\\
24 & 61.3$\pm$4.6 \\
28 & 62.2 $\pm$4.8\\
32 & 61.7$\pm$5.1 \\
\hline
\end{tabular}
\caption{Performance vriation with reasoning time.}
\label{tab:placeholder}
\end{table}

%%%%%% add qualitative examples here as well of the trajectory - no need to add in main unless comments suggest otherwise
%%%% qualitative examples need to show some steps in the trajectory
% 

% For the prostate cancer segmentation task, we also present intermediate reasoning steps i.e., winning refinements from intermediate ...

% \begin{figure}
%     \centering
%     \includegraphics[width=0.5\linewidth]{figs/intermediate_steps_prostate.png}
%     \caption{Caption}
%     \label{fig:intermediate_steps}
% \end{figure}

\paragraph*{Computational cost}

The computational cost of the GenVerOpt framework is summarised in Tab. \ref{tab:carbon}. Note that the time quoted for the `Reason' phase is for a single sample. In practice a batch of 24 samples can be processed in parallel on the described hardware.

\begin{table}[!ht]
\centering
\begin{tabular}{|c| c| c |c |c |c|}
% Hardware & GPU Hours (pre-train, adapt, infer - (0 experience, 8 experience, 12 experience,)) & Power & Carbon \\
\hline
Hardware & \multicolumn{3}{|c|}{GPU Hours} & Power (W) & Carbon (kg CO2/h)\\ 
\hline
& Pre-train & Adapt & Reason & &\\
\hline
Nvidia A100 & 672 & 16 & 26 & 300W & 0.13\\
\hline
\end{tabular}
\caption{Carbon footprint.}
\label{tab:carbon}
\end{table}

\bibliography{sn-bibliography}% common bib file

\end{document}